\documentclass{article}
\usepackage{amsmath,graphicx,mlspconf}
\usepackage{xcolor}

\usepackage{cite}
\usepackage{graphicx}
\usepackage{amsmath}
\usepackage{wasysym}
\usepackage{amssymb}
\usepackage{subcaption}
\usepackage{multicol}
\usepackage{tabularx}
\usepackage{algorithmic}
\usepackage{diagbox}

\usepackage{array}
\usepackage{bm} 
\usepackage{xcolor}
\usepackage{fixltx2e}

\usepackage{multirow}
\usepackage{dblfloatfix}
\usepackage{algorithm}

\usepackage{graphicx}
\usepackage{booktabs}
\usepackage{colortbl}
\usepackage[table,xcdraw]{xcolor}
\usepackage{float}

\usepackage{caption}
\usepackage[top=0.75in, bottom=0.75in, left=0.7in, right=0.7in]{geometry}

\usepackage{url}
\usepackage{makecell}
\usepackage{hyperref}

\usepackage[dvipsnames]{xcolor}

\copyrightnotice{979-8-3503-2411-2/25/\$31.00 {\copyright}2025 Crown}

\toappear{\tiny 2025 IEEE International Workshop on Machine Learning for Signal Processing, Aug.\ 31-- Sep.\ 3, 2025, Istanbul, Turkey}

\title{Compression Beyond Pixels:\  \ Semantic Compression with Multimodal Foundation Models}
\name{Ruiqi Shen, Haotian Wu, Wenjing Zhang, Jiangjing Hu, Deniz Gunduz\thanks{This work was supported by UKRI under the projects AI-R (EP/X030806/1) and INFORMED-AI (EP/Y028732/1), and by the SNS JU project 6G-GOALS under the EU Horizon program (Grant Agreement No. 101139232). Corresponding author: Haotian Wu (haotian.wu17@imperial.ac.uk).}}
\address{Department of Electrical and Electronic Engineering, Imperial College London}

\begin{document}
\maketitle

\begin{abstract}
Recent deep learning-based methods for lossy image compression achieve competitive rate-distortion performance through extensive end-to-end training and advanced architectures. However, emerging applications increasingly prioritize semantic preservation over pixel-level reconstruction and demand robust performance across diverse data distributions and downstream tasks. These challenges call for advanced semantic compression paradigms. Motivated by the zero-shot and representational capabilities of multimodal foundation models, we propose a novel semantic compression method based on the contrastive language-image pretraining (CLIP) model. Rather than compressing images for reconstruction, we propose compressing the CLIP features into minimal bits while preserving semantic information across different tasks. Experiments show that our method maintains semantic integrity across benchmark datasets, achieving an average bit rate of approximately $2\text{--}3 \times 10^{-3}$ bits per pixel. This is less than $5\%$ of the bitrate required by mainstream image compression approaches for comparable performance. Remarkably, even under extreme compression, the proposed approach exhibits zero-shot robustness across diverse data distributions and downstream tasks. Code and checkpoints are available at \url{https://github.com/JasonShen-SH/CLIP_compress}.
\end{abstract}


\section{Introduction}
Deep learning has revolutionized lossy image compression by introducing data-driven codecs that replace classical analysis and synthesis transforms with neural networks. 
Starting with the adoption of autoencoders and convolutional neural networks, neural image compression has progressed through several key innovations, including the integration of hyperpriors \cite{balle2018variational}, attention mechanisms \cite{cheng2020learned}, and advanced overfitted coding \cite{wulotterycodec}. These developments have substantially improved the rate-distortion (RD) performance, reshaping the landscape of lossy image compression.

Concurrently, emerging applications from autonomous systems \cite{mao2024diffcp} to edge intelligence \cite{hua2023edge} increasingly prioritize semantic processing for downstream vision tasks over full image reconstruction \cite{hu2025task}. These often involve real-time decision-making under strict delay and resource constraints. This shift has motivated task-oriented semantic compression methods \cite{gunduz2022beyond} that focus on retaining key features for the task at hand, such as classification \cite{luo2018deepsic} or object detection \cite{mao2024diffcp,wang2023semantic}. However, these approaches assume the task is known, and require retraining for new tasks, limiting their flexibility and scalability in dynamic environments. Leveraging recent breakthroughs in multimodal foundation models (MFMs) \cite{radford2021learning,liu2023visual}, we propose a reconstruction-free semantic compression paradigm that directly compresses MFM-generated token embeddings, robustly preserving essential semantics for downstream tasks. Specifically, we employ the contrastive language-image pretraining (CLIP) model \cite{radford2021learning} to generate compact image embeddings (referred to as CLIP features), which are then compressed into bit-level representations. To enable this, we introduce a product-quantization-based variational autoencoder with a shared codebook (\textit{PQVAE-shared}), which compresses CLIP features into the minimum number of bits required to retain semantic information. This task-agnostic approach achieves superior compression ratios while maintaining robust generalization across diverse data distributions and downstream tasks, outperforming both pixel-level image compression and task-specific semantic compression techniques.

Our key contributions can be summarized as follows:
\begin{itemize}
    \setlength{\itemsep}{0pt}
    \item We propose CLIP-driven semantic compression, a novel paradigm that preserves essential semantic information at extremely low bit rates.
    \item We introduce \textit{PQVAE-shared}, a product-quantization-based variational autoencoder with a shared codebook to compress CLIP features. It mitigates codebook collapse by employing a more compact and efficiently utilized codebook.
    \item Comprehensive experiments against learned image compression methods demonstrate on-par or superior performance while requiring less than 5\% of their bit rate $(2\text{--}3 \times 10^{-3}$ bits per pixel on standard benchmarks).
    \item The proposed \textit{PQVAE-shared} scheme shows robust zero-shot performance across diverse data distributions and downstream tasks, even under extreme rates.
\end{itemize}

\section{System Model}
\begin{figure}[t]
    \centering
    \includegraphics[width=1\linewidth]{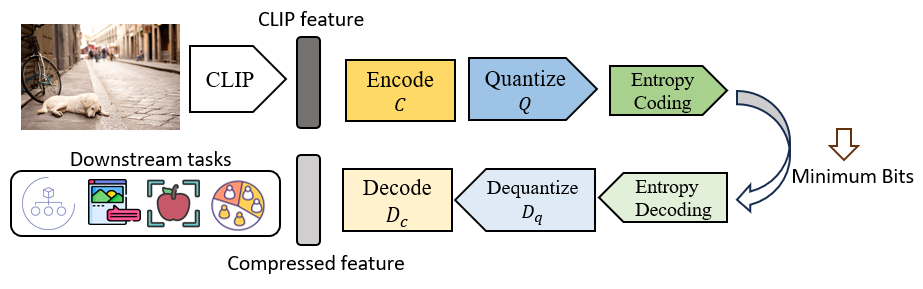}
    \caption{Overview of MFM-driven semantic compression.}
    \label{fig:Overall method}
\end{figure}

Our goal is to design a robust semantic compression scheme for various downstream tasks, without requiring the reconstruction of the original image. In this paper, we exploit the CLIP model \cite{radford2021learning}, which aligns images’ visual features with corresponding textual descriptions using contrastive learning. Our focus is on compressing CLIP features to the minimal number of bits while preserving robust semantic information of the corresponding images, making this approach suitable for rate-limited edge applications. See Fig.~\ref{fig:Overall method} for the illustration of the proposed CLIP-driven semantic compression framework.

\subsection{Compressor} 
Given an image \( \mathbf{I} \in \mathbb{R}^{H_0 \times W_0 \times C_0} \), where \( H_0 \), \( W_0 \), and \( C_0 \) are the height, width, and number of channels of the image, respectively, the CLIP image encoder \( F(\cdot) \) maps it to a \( k \)-dimensional embedding space, \(\mathbb{R}^k\), i.e., \( F(\cdot): \mathbb{R}^{H_0 \times W_0 \times C_0} \rightarrow \mathbb{R}^k \). Let \( \mathbf{x} = F(\mathbf{I}) \), to compress \( \mathbf{x} \in \mathbb{R}^{k} \), we first transform it into a latent representation \( \mathbf{x}_c \in \mathbb{R}^{h \times w \times c} \) via an encoder network \( \mathcal{C}(\cdot): \mathbb{R}^k \rightarrow \mathbb{R}^{h \times w \times c} \), given as: 
\begin{equation}
\vspace{-3pt}
\mathbf{x}_c = \mathcal{C}(\mathbf{x}),
\label{eq:compressor_encoding}
\vspace{-3pt}
\end{equation}
where $h$ and $w$ correspond to the spatial dimensions, while $c$ represents the fixed channel dimension at each spatial location.

Subsequently, the quantization process \( Q(\cdot) : \mathbb{R}^{h \times w \times c} \to \mathcal{Z}^{d_q} \) is applied to \( \mathbf{x}_c \) as: 
\begin{equation}
    \vspace{-3pt}
    \mathbf{z}_q = Q(\mathbf{x}_c),
    \vspace{-3pt}
\end{equation}
mapping \( \mathbf{x}_c \) to the corresponding integer indices of the codebook.
To further compress \( \mathbf{z}_q \), an entropy coding method is employed, with its results contributing to the final bit stream. 

\subsection{Decompressor} 
At the decoder, the bitstream undergoes entropy decoding to recover \( \mathbf{{z}}_q \), which is then de-quantized via \( \mathcal{D}_Q(\cdot) : \mathcal{Z}^{d_q} \to \mathbb{R}^{h \times w \times c} \), for reconstructing \({\mathbf{x}}_c\) as \(\hat{\mathbf{x}}_c\):
\begin{equation}
    \vspace{-3pt}
    \hat{\mathbf{x}}_c =\mathcal{D}_Q(\mathbf{\hat{z}}_q).
    \vspace{-3pt}
\end{equation}
Similarly, a decompression process \( \mathcal{D}_C(\cdot) : \mathbb{R}^{h \times w \times c} \to \mathbb{R}^k \) recovers the MFM feature as \( \hat{\mathbf{x}} \):
\begin{equation}
\vspace{-3pt}
\hat{\mathbf{x}} = \mathcal{D}_C(\hat{\mathbf{x}}_c).
\vspace{1pt}
\end{equation}

\subsection{R-D trade-off}
We minimize the compression rate $R$ while constraining the semantic distortion $D$. More formally, we quantify $R$ using the number of bits per feature dimension (bpd), which refers to the average number of bits required to encode each dimension of the CLIP feature vector. Distortion \( D \) is defined as the semantic inconsistency using cosine similarity:
\begin{equation}
    D(\mathbf{x}, \hat{\mathbf{x}}) \triangleq 1 - \frac{\hat{\mathbf{x}} \cdot \mathbf{x}}{\|\hat{\mathbf{x}}\| \|\mathbf{x}\|},
\end{equation}
where $||\mathbf{x}||$ denotes the Euclidean norm of $\mathbf{x}$.

To minimize $R$ while keeping $D$ within an acceptable threshold, the problem is formulated as:
\begin{equation}
\vspace{-3pt}
\begin{array}{cl}
 \textbf{P1:}& \min\limits_{\mathcal{C},{Q},\mathcal{D}_Q,\mathcal{D}_C} \mathbb{E} \left[R\right],\quad \text{subject to }\mathbb{E} \left[D\right] \leq D_{\text{max}},  
\end{array}
\vspace{-3pt}
\label{eq_p2}
\end{equation}
where $D_{\text{max}}$ is the maximum allowable semantic distortion, and the expectation is computed across the dataset.

Since CLIP image encoder is trained using 32-bit float representation, the original CLIP embedding can be conveyed as is (i.e., without any compression) using 32 bpd. However, we will show that it is possible to further compress the embedding to significantly lower bpd values (less than 1 bpd) without significantly hurting the downstream task performance, while maintaining generalizability across diverse data distributions.

\section{Product Quantization-Variational Autoencoder With Shared Codebook}

\subsection{Overall architecture}

\begin{figure}[t]
    \centering
\includegraphics[width=1.0\linewidth]{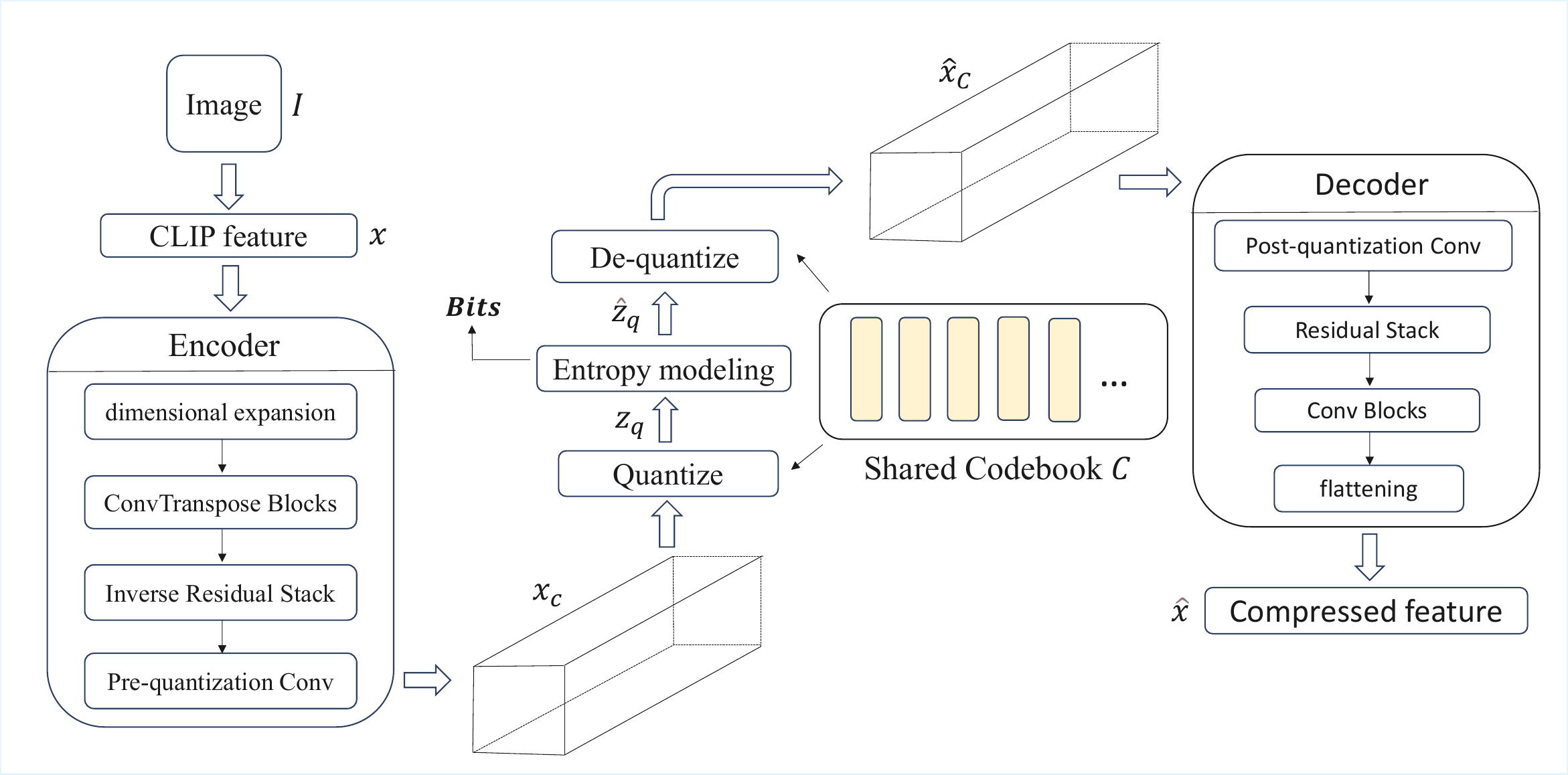}
    \caption{Overview of the proposed \textit{PQVAE-shared} scheme.}
    \label{fig:pqvae framework}
\end{figure}

The proposed \textit{PQVAE-shared} scheme for feature compression is illustrated in Fig. \ref{fig:pqvae framework}, which employs a learned product quantization module and a shared codebook.

After obtaining the latent representation \( \mathbf{x_c} \in \mathbb{R}^{h \times w \times c} \) as in Eq.~\eqref{eq:compressor_encoding}, a product quantization method with a shared codebook \( \mathbf{C} \) is independently applied to each channel of \( \mathbf{x_c}\). Each channel of dimension \( c \) is first decomposed into \( d \) lower-dimensional subspaces, and each component is mapped to a codeword from \( \mathbf{C} \), resulting in an index matrix \( \mathbf{z}_q \in \mathbb{Z}^{h \times w \times d} \). To further minimize the bitstream size and to remove any remaining redundancy, entropy coding is employed over $\mathbf{z}_q$. For decompression, the bitstream is processed to retrieve the integer indices $\mathbf{\hat{z}}_q \in \mathbb{Z}^{h \times w \times d}$, which are then de-quantized to recover $\hat{\mathbf{x}}_c$ and subsequently decoded to reconstruct the estimated feature embedding $\hat{\mathbf{x}}$.


\subsection{Product quantization with shared codebook} \label{sec:pq_shared_codebook}

\begin{figure}[t]
    \centering
    \includegraphics[width=0.89\linewidth]{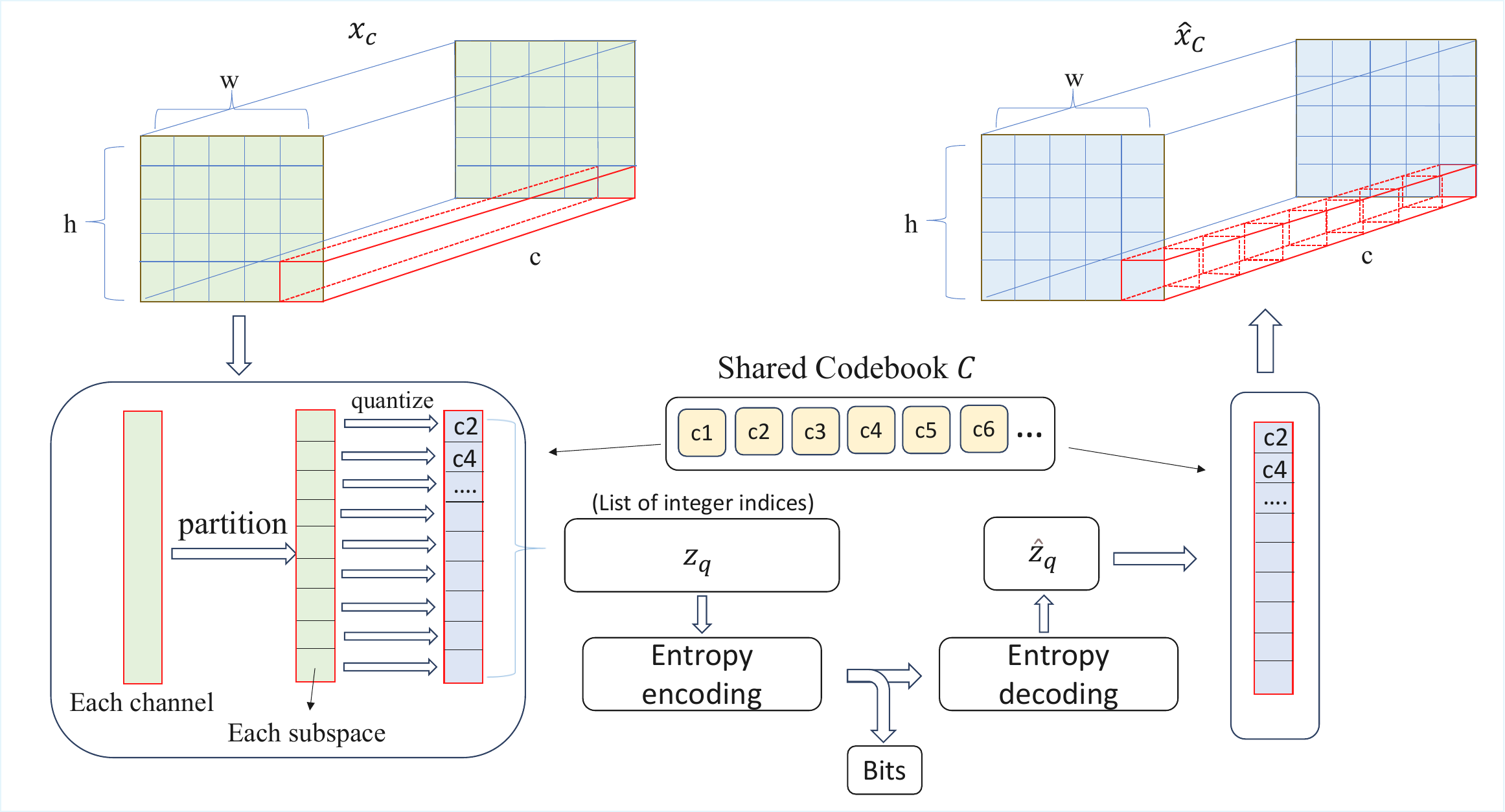}
    \caption{Product quantization with shared codebook.}
    \label{fig:pqvae quantization}
\end{figure}

As shown in Fig. \ref{fig:pqvae quantization}, for each channel $\mathbf{s}_i \in \mathbb{R}^c$ (\( i \in \{1, \dots, h \times w\} \)) of $\mathbf{x}_c$, quantization begins by dividing it into \( d \) lower-dimensional subvectors within \( d \) corresponding subspaces, each of fixed size $d_{\text{sub}} = \frac{c}{d}$. The shared codebook $\mathbf{C}$ is then applied to quantize the subspaces' embeddings into integer indices via nearest-neighbor search, each index corresponding to a codeword. The $d$ integer indices are subsequently entropy encoded into a bitstream. Upon decoding, the indices are recovered, and the corresponding codewords are retrieved from $\mathbf{C}$ to reconstruct $\mathbf{s}_i$ as $\mathbf{\hat{s}}_i$. This process is applied to all $h \times w$ channels, with quantized representations combined to form $\hat{\mathbf{x}}_c \in \mathbb{R}^{h \times w \times c}$. See Algorithm \ref{alg:shared_codebook_pq} for details.

The compression cost is determined by the total bit representation of the indices, which is \(h \times w \times d \times \lceil \log_2(K) \rceil\), where \(K\) is the size of codebook \(\mathbf{C}\).  This is then further compressed by entropy coding, yielding the final bit cost. 

\subsection{Training strategy} \label{sec:training}
The overall loss for training \textit{PQVAE-shared} scheme is: 
\begin{equation}
    L = \left(1 - \frac{\hat{\mathbf{x}} \cdot \mathbf{x}}{\|\hat{\mathbf{x}}\| \|\mathbf{x}\|}\right) + \alpha \| s_g(\mathbf{x}_c) - \hat{\mathbf{x}}_c \|^2 + \beta  \| \mathbf{x}_c - s_g(\hat{\mathbf{x}}_c) \|^2,
    \label{loss}
\end{equation}
where the first term is the cosine similarity loss, measuring semantic distortion during compression. The second and third loss terms are the codebook loss and commitment loss, adopted from \cite{van2017neural}, which optimize the discrete codebook representations. $\alpha$ and $\beta$ are hyperparameter that balance different loss terms. Specifically, the codebook loss pulls every codeword closer to the encoder output it represents, while the commitment loss pushes the encoder output to stay near its chosen codeword and keeps its scale in check. $s_g(\cdot)$ stands for the stop gradient operator. 

\section{Experimental Results}
\subsection{Experiment setup} 
Our compression framework of \textit{PQVAE-shared} is trained on ImageNet, using 800 classes for training and the other 200 classes for validation. \textcolor{black}{To evaluate semantic quality of feature reconstruction, we conduct zero-shot classification, image captioning, and referring object identification at the receiver.} Specifically, we employ the pre-trained CLIP model \textit{``ViT-L/14@336px"} to encode images into $768$-dimensional embeddings, which are then compressed with our module. Huffman coding is employed for entropy coding throughout the experiments. 

\textcolor{black}{For baselines, we employ \textit{VQ-VAE} scheme with k-means product quantization~\cite{wu2019learning,van2017neural} and learned scalar quantization (SQ) \footnote{Learned scalar quantization learns a set of optimal quantization levels between -0.5 and 0.5 for minimal quantization error.}, as they closely resemble our framework and target the semantic compression of CLIP features. Additional competitive image compression methods are also included for comparison. We note that other baselines, such as task-specific semantic compression approaches~\cite{luo2018deepsic, wang2023semantic, shindo2024image} that do not leverage foundation models, exhibit limited generalization and perform significantly worse in zero-shot experiments.}

\begin{algorithm}[t]
\caption{Product quantization using shared codebook}
\label{alg:shared_codebook_pq}
\textbf{Input:} $\mathbf{x}_c \in \mathbb{R}^{h \times w \times c}$,  codebook $\mathcal{C}$, entropy dictionary $\mathcal{H}$ \\
\textbf{Output:} bitstream $\mathbf{b}$, $\mathbf{\hat{x}}_c \in \mathbb{R}^{h \times w \times c}$

\begin{algorithmic}[1]
\FOR{each channel $\mathbf{s}_{i} \in\mathbb{R}^{c}$ from $\mathbf{s}_1$ to $\mathbf{s}_{hw}$}
    \STATE Segment $\mathbf{s}_{i}$ into $d$ subspaces: $\mathbf{s}_{i1}, \ldots, \mathbf{s}_{id}$.
    \FOR{each subspace \( \mathbf{s}_{ij} \in \mathbb{R}^{c_{\text{sub}}} \)}
        \STATE Quantize $\mathbf{s}_{ij}$ to $\mathbf{c}_k\in \mathcal{C}$ with the index $k$ as $\mathbf{z}_{ij}$
    \ENDFOR
    \STATE Form index vector $\mathbf{z}_i = \{\mathbf{z}_{i1}, \ldots, \mathbf{z}_{id}\}$.
    \STATE Entropy encoding using $\mathcal{H}$: $\mathbf{z}_i\rightarrow \mathbf{b}$.
    \STATE Entropy decoding using $\mathcal{H}$: $\mathbf{b}\rightarrow \mathbf{\hat{z}}_i$
    \STATE Dequantize \( \mathbf{\hat{z}}_i \) back to codewords, yielding \( \mathbf{\hat{s}}_i \in \mathbb{R}^{c} \).
\ENDFOR
\STATE Reconstruct $\mathbf{\hat{x}}_c \in \mathbb{R}^{h \times w \times c}$ from $\mathbf{\hat{s}}_1$ to $\mathbf{\hat{s}}_{hw}$.
\end{algorithmic}
\end{algorithm}

\begin{figure}[t]
    \centering
    \includegraphics[width=1.0\linewidth]{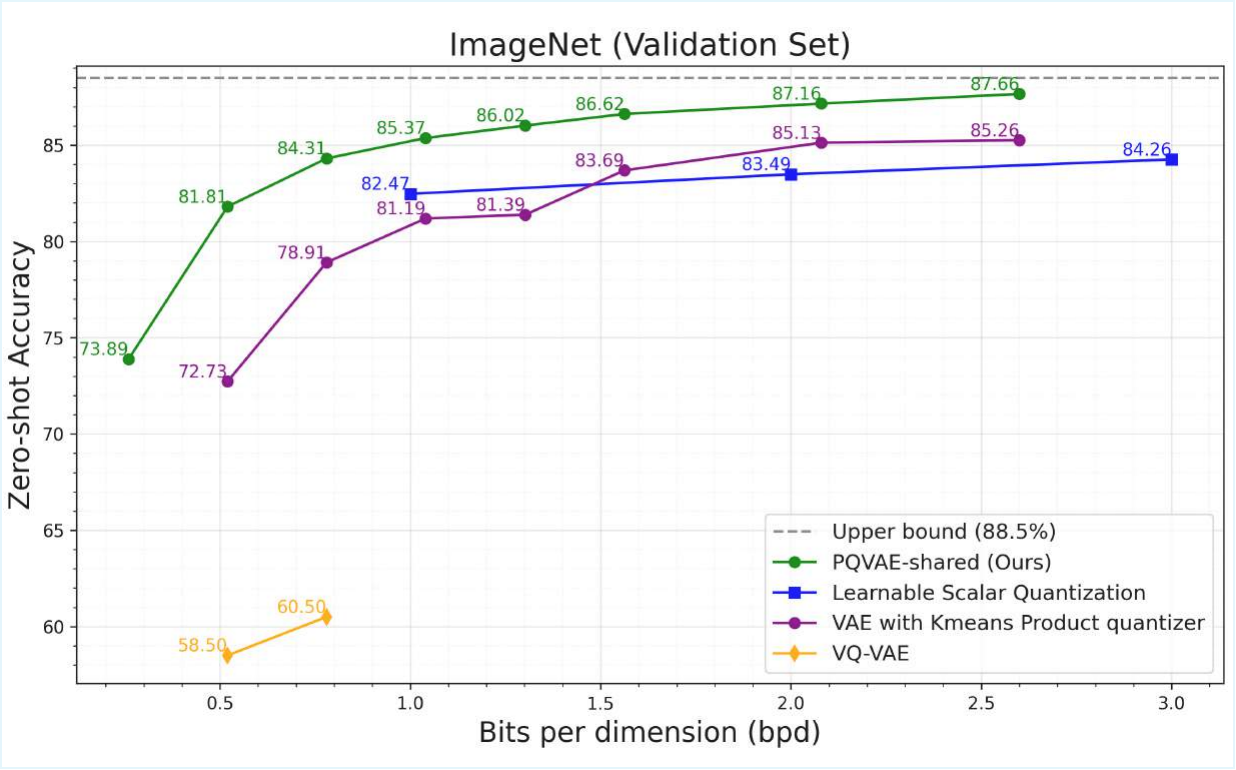}
    \caption{Comparison of \textit{PQVAE-shared} and baselines for CLIP-driven semantic compression on ImageNet validation set.}
    \label{fig:pqvae result}
\end{figure}

\subsection{General performance}


{As shown in Fig.~\ref{fig:pqvae result}, \textit{PQVAE-shared} consistently outperforms all baselines in zero-shot classification across every bpd. Here, the “upper bound” refers to zero-shot accuracy using the original CLIP features (32 bpd). Even at \(0.52\) bpd (400 bits/image), our method achieves 81.81 \%, over 30× compression with only a small accuracy drop. As the bit budget increases, performance smoothly rises toward the upper bound.}

{Standard VQ-VAE allocates a distinct codeword to each spatial location such that under a total bit budget \(B\), each \(c\)-dimensional vector at one of the \(H\times W\) positions could be encoded with \(B/(H\times W)\) bits, yielding a codebook of size \(2^{B/(H\times W)}\), which becomes infeasible when \(B\) or \(H\times W\) is large. For example, suppose \(H=W=5\), \(c=128\) (a regular codebook size in VQ-VAE), and a bit budget of \(B=600\) bits (corresponding to 0.78 bpd), then each 128-dim vector would require \(600/(5\times5)=24\) bits—i.e.\ a codebook of size \(2^{24}\approx1.68\times10^{7}\), far beyond standard hardware capacity. By contrast, \textit{PQVAE-shared} splits each 128-dim channel into \(d\) equal subspaces, so that each subvector is encoded with \(600/(5\times5\times d)\) bits, requiring a shared codebook of size \(2^{600/(5\times5\times d)}\). For example, setting \(d=8\) yields \(600/(5\times5\times8)=3\) bits per subvector and a codebook of size \(2^3=8\), thus enabling practical training and inference while preserving semantic fidelity, achieving 84.31\% zero-shot classification accuracy (Fig.~\ref{fig:pqvae result}). Meanwhile, learned scalar quantization of CLIP features also delivers competitive results but its bpd values are restricted to discrete integers, preventing the fine-grained bit control like our approach.}

\subsection{Ablation study on subspace decomposition}
\begin{figure}[t]
    \centering
    \includegraphics[width=0.9\linewidth]{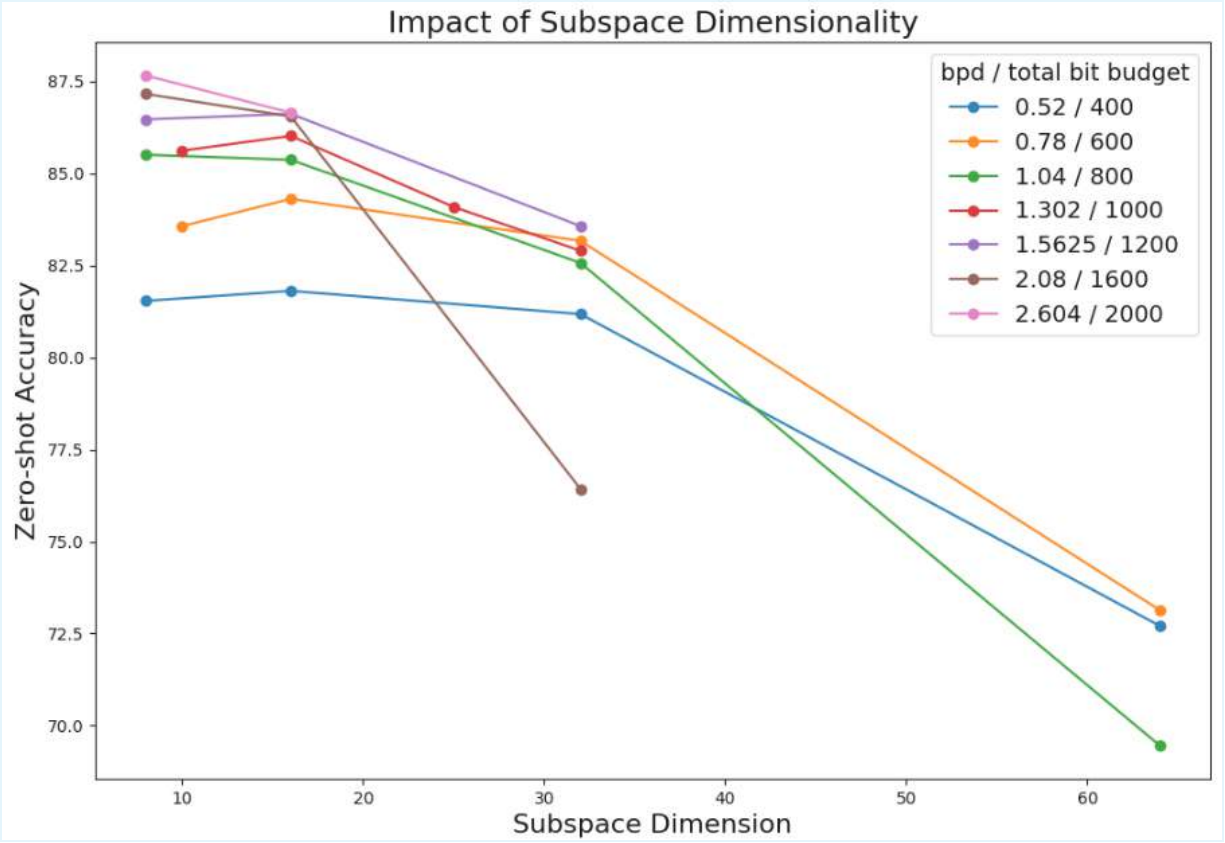}
    \caption{Impact of varying subspace dimensions at different bpd values on the ImageNet validation set.}
    \label{fig:ablation study}
\end{figure}

To study the impact of subspace decomposition, we fix $\mathbf{x}_c$ with dimension \( 5 \times 5 \times 128 \)\footnote{\textcolor{black}{The latent space dimensionality is intentionally increased ({exceeding the standard CLIP embedding dimension of 768}) to facilitate more effective codebook learning in product quantization.}}, and vary the subspace. As shown in Fig. \ref{fig:ablation study}, finer subspace decomposition obtained by reducing \( d_{\text{sub}} \) consistently improves performance at all bpd levels by preserving more semantic details and using the codebook more efficiently. However, performance might also drop slightly when \( d_{\text{sub}} \) becomes too small due to limited codebook capacity. These results highlight the need to balance decomposition granularity and codebook capacity for optimal performance.

\vspace{-3mm} 

\subsection{Comparison to image compression methods}
\begin{figure*}[t]
    \centering
    \begin{subfigure}{0.49\textwidth}
        \centering
        \includegraphics[width=1\linewidth]{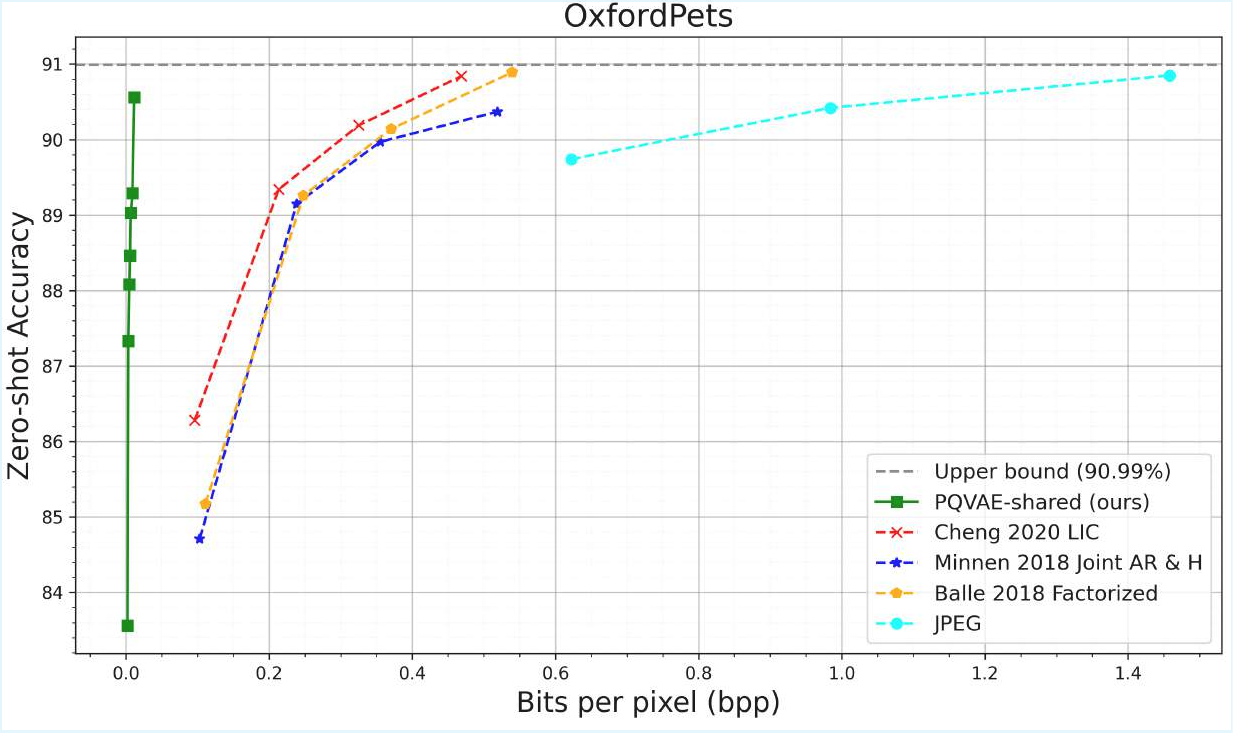}
        \caption{OxfordPets}
        \label{fig:oxford}
    \end{subfigure}
    \hfill
    \begin{subfigure}{0.49\textwidth}
        \centering
        \includegraphics[width=1\linewidth]{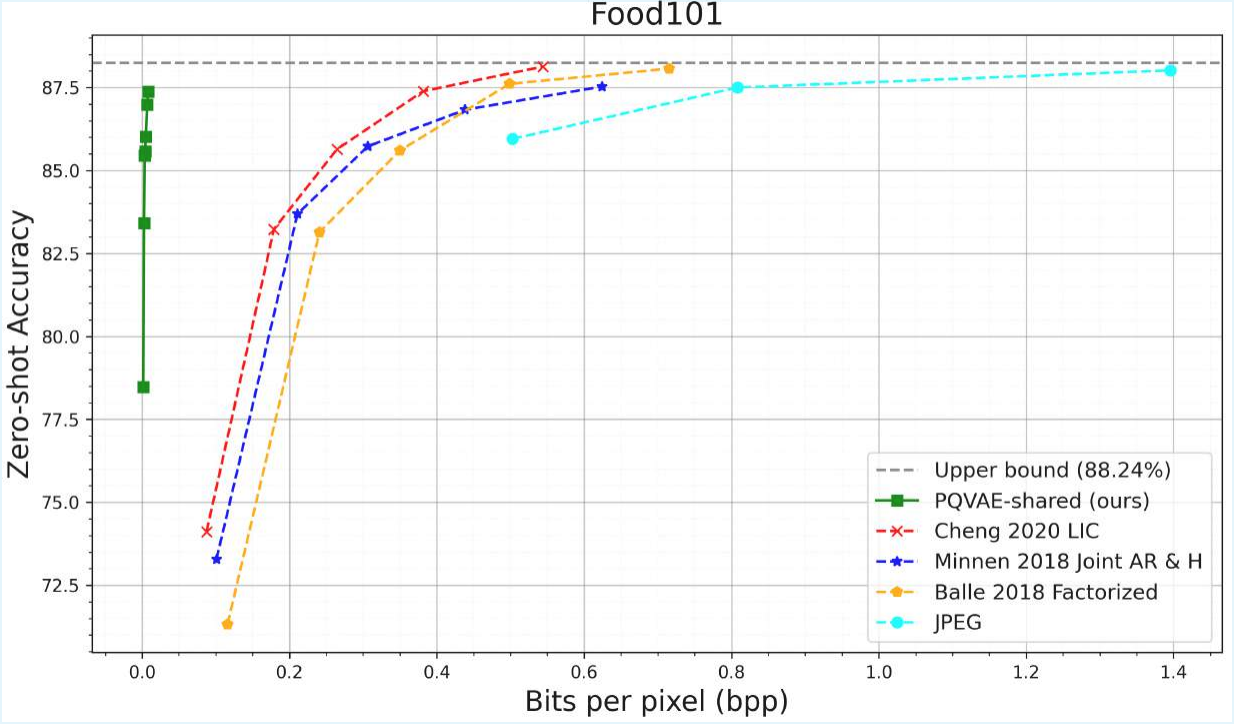}
        \caption{Food101}
        \label{fig:food}
    \end{subfigure}
    \caption{Comparison of zero-shot classification performance between \textit{PQVAE-shared} and image compression baselines.}
    \label{fig:comparison with image compression}
\end{figure*}

We compare the proposed \textit{PQVAE-shared} with competitive image compression methods on OxfordPets and Food101 datasets. For fair comparison, compression levels here are measured in bits per pixel (bpp), and semantic preservation is evaluated using zero-shot image classification accuracy. As shown in Fig. \ref{fig:comparison with image compression}, on OxfordPets, our method achieves accuracies of 83.56\% at \(2.29 \times 10^{-3}\) bpp and 87.33\% at \(3.43 \times 10^{-3}\) bpp, using less than 3\% of the bpp required by learned image compression methods like \textit{Cheng2020-anchor} \cite{cheng2020learned}, which achieves 86.28\% accuracy with \(96.1 \times 10^{-3}\) bpp. Similarly, on Food101, our method achieves 78.47\% accuracy at \(1.723 \times 10^{-3}\) bpp, surpassing \cite{cheng2020learned} for 74.11\% accuracy under \(87.5 \times 10^{-3}\) bpp. This demonstrates our method’s efficiency in preserving semantics using only 2\text{--}3\% of the typical bpp of learned compression approaches, achieving extreme compression without compromising effectiveness.

\vspace{-3mm} 

\subsection{Zero-shot performance evaluation}
We further assess the robustness of our method by evaluating its zero-shot performance across multiple benchmark datasets and comparing it to learned scalar quantization (SQ). As shown in Table \ref{tab:zero_shot_performance}, our experiments span four benchmark datasets with bpd values ranging from 0.52 bpd (400 bits) to 3 bpd (2304 bits). 
\textit{PQVAE-shared} consistently outperforms learned SQ across all bpd values, achieving nearly identical accuracy to the upper bound while maintaining 10x higher compression ratio (2.604 bpd vs. 32bpd). 

\begin{table*}[t]
\centering
\caption{Comparison of different schemes at various bpd values across multiple datasets. \textit{UB} denotes the upper bound accuracy.}
\centering
\renewcommand{\arraystretch}{1.2}
\setlength{\tabcolsep}{8pt}
\begin{tabular}{|l|l|c|c|c|c|c|c|c|c|c|}
\hline
\textbf{Dataset} & \textbf{method / bpd} & \textbf{0.52} & \textbf{0.78} & \textbf{1} & \textbf{1.04} & \textbf{1.3} & \textbf{1.56} & \textbf{2} & \textbf{2.604} & \textbf{3} \\ \hline
\multirow{2}{*}{\makecell[l]{OxfordPets \\ (\textit{UB}:90.99)}} & \textit{PQVAE-shared} & 83.55 & 87.30 & - & 88.08 & 88.46 & 89.15 & - & 90.52 & -  \\ \cline{2-11}
                            & Learned SQ & - & - & 81.33 & - & - & - & 82.28 & - & 83.56   \\ \hline
\multirow{2}{*}{\makecell[l]{Caltech101 \\ (\textit{UB}:88.84)}} & \textit{PQVAE-shared} & 85.43 & 86.12 & - & 86.91 & 87.60 & 88.03 & - & 88.07 & - \\ \cline{2-11}
                            & Learned SQ & - & - & 83.28 & - & - & - & 83.38 & - & 83.61   \\ \hline
\multirow{2}{*}{\makecell[l]{Tiny-ImageNet \\ (\textit{UB}:71.73)}} & \textit{PQVAE-shared} & 63.71 & 66.95 & - & 67.78 & 68.10 & 68.94 & - & 70.62 & - \\ \cline{2-11}
                              & Learned SQ & - & - & 62.99 & - & - & - & 66.27 & - & 67.84   \\ \hline
\multirow{2}{*}{\makecell[l]{Food101 \\ (\textit{UB}:88.84)}} & \textit{PQVAE-shared} & 79.39 & 84.48 & - & 85.99 & 86.16 & 87.00 & - & 87.98 & - \\ \cline{2-11}
                            & Learned SQ & - & - & 84.22 & - & - & - & 84.88 & - & 85.58 \\ \hline
\end{tabular}
\label{tab:zero_shot_performance}
\end{table*}

\subsection{Performance on downstream tasks}
To ensure compressed CLIP features maintain semantic integrity for robust downstream performance, we further evaluate our approach on two representative tasks: image captioning and referring object identification, both with the pretrained feature compression framework of \textit{PQVAE-shared} frozen.

\begin{figure*}[t]
    \centering
    \begin{subfigure}{0.49\textwidth}
\includegraphics[width=1\linewidth, height=5cm]{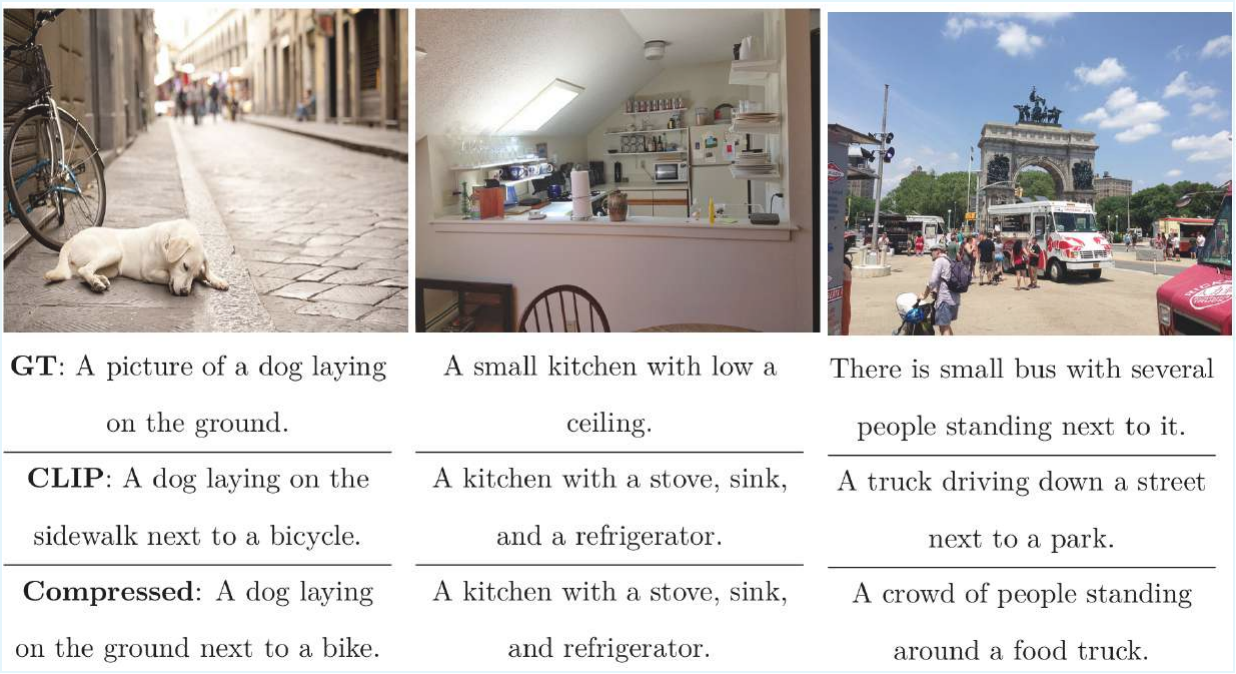}
    \end{subfigure}
    \begin{subfigure}{0.49\textwidth}
\includegraphics[width=1\linewidth, height=5cm]{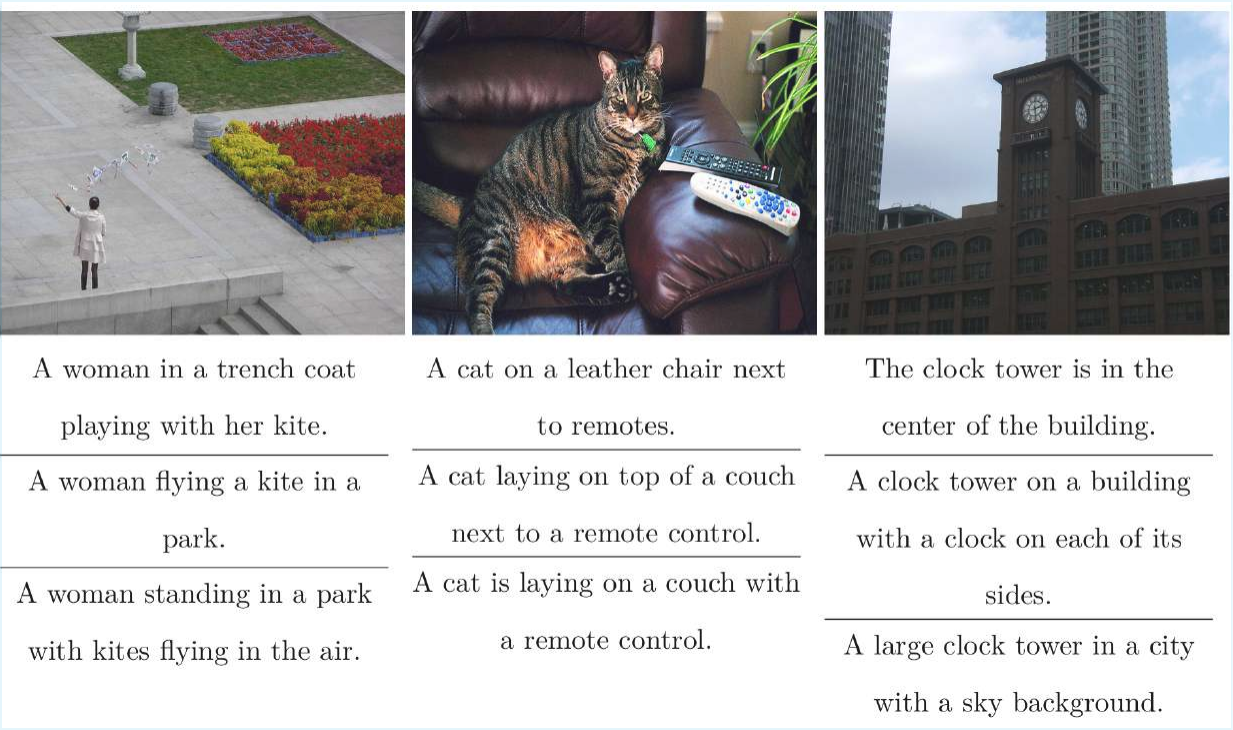}
    \end{subfigure}
    \caption{Image captioning results. Top row: ground truth (\textit{GT}); middle row: captions from original CLIP features (\textit{CLIP}); bottom row: captions from compressed CLIP features (\textit{Compressed}), with each image compressed to only 400 bits.}
    \label{fig:real_caption}
\end{figure*}

\textbf{\textit{(a) Image captioning.}} We extend ClipCap \cite{mokady2021clipcap} by inserting the compression module between the CLIP image encoder and mapping network, and fine-tuning only the mapping network and GPT-2—exactly the same method as in \cite{mokady2021clipcap}. Evaluated at 400 bits per image on COCO2014, the compressed features preserve semantic integrity by generating captions nearly identical to those from original CLIP features (See Fig. \ref{fig:real_caption}).

\begin{figure*}[t]
  \centering 
  \includegraphics[width=1\textwidth]{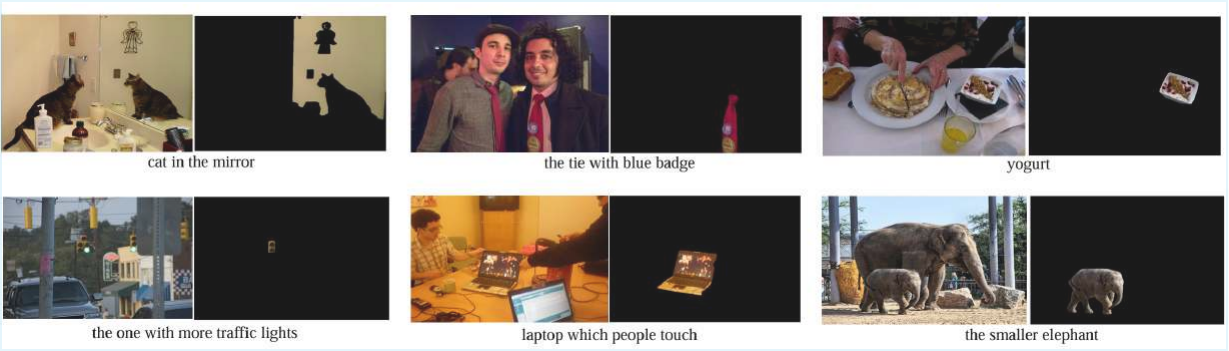}
  \caption{Referring object identification results, with each image compressed to only 400 bits.}
  \label{fig: Real sam pics}
\end{figure*}

\textbf{\textit{(b) Referring object identification.}} This task localizes objects using text descriptions in a plug-and-play manner. First, Segment Anything Model(SAM)\cite{kirillov2023segment} segments the image into candidate regions, which are then encoded into CLIP features and compressed using \textit{PQVAE-shared}. The compressed features are compared with CLIP-encoded text embeddings via cosine similarity to identify the best-matching region. This ensures privacy protection by preventing visual reconstruction, and seamless integration, as the selected region’s CLIP feature retains semantics ready for downstream tasks. As shown in Fig.~\ref{fig: Real sam pics}, robust accuracy is maintained even under extreme compression (400 bits per image) with challenging prompts.

\section{Conclusion}
We presented CLIP-driven semantic image compression, implemented through product quantization-variational autoencoder with shared codebook, which is task-agnostic and focuses on semantic preservation for downstream tasks rather than pixel-level reconstruction. Experiments demonstrate that original CLIP features can be compressed more than 30-fold while maintaining satisfactory semantic preservation. Compared with other learned image compression methods, our approach achieves a significantly lower bitrate (averaging $2\text{--}3 \times 10^{-3}$ bits per pixel) across standard benchmark datasets, while maintaining robustness across diverse data distributions and downstream tasks. The proposed paradigm is adaptable to a wide range of MFMs, enabling more efficient and versatile semantic compression for future real-world applications.





\bibliographystyle{IEEEbib}
\bibliography{MLSP}

\end{document}